\documentclass[10pt,twocolumn,letterpaper]{article}

\usepackage{wacv}              
\usepackage{graphicx}
\usepackage{amsmath}
\usepackage{amssymb}
\usepackage{booktabs}
\usepackage[table, dvipsnames]{xcolor}

\usepackage[pagebackref,breaklinks,colorlinks]{hyperref}

\usepackage[capitalize]{cleveref}
\crefname{section}{Sec.}{Secs.}
\Crefname{section}{Section}{Sections}
\Crefname{table}{Table}{Tables}
\crefname{table}{Tab.}{Tabs.}

\def\wacvPaperID{956} 
\def\confName{WACV}
\def\confYear{2025}

\newcommand{\myparagraph}[1]{\vspace{2pt}\noindent\textbf{#1}}

\def\newTaskName{Negative-Aware Video Moment Retrieval}
\def\newTaskNameShort{NA-VMR}
\def\methodName{UniVTG-NA}

\begin{document}

\title{Moment of Untruth: Dealing with Negative Queries in Video Moment Retrieval}

\author{Kevin Flanagan\\
University of Bristol\\
{\tt\small kevin.flanagan@bristol.ac.uk}
\and
Dima Damen\\
University of Bristol \\
{\tt\small dima.damen@bristol.ac.uk}
\and
Michael Wray\\
University of Bristol \\
{\tt\small michael.wray@bristol.ac.uk}
}
\maketitle

\begin{abstract}
Video Moment Retrieval is a common task to evaluate the performance of visual-language models---it involves localising start and end times of moments in videos from query sentences. 
The current task formulation assumes that the queried moment is present in the video, resulting in false positive moment predictions when irrelevant query sentences are provided.

In this paper we propose the task of \newTaskName{} (\newTaskNameShort), which considers both moment retrieval accuracy and negative query rejection accuracy.
We make the distinction between In-Domain and Out-of-Domain negative queries and provide new evaluation benchmarks for two popular video moment retrieval datasets: QVHighlights and Charades-STA. 
We analyse the ability of current SOTA video moment retrieval approaches to adapt to \newTaskName{} and propose \methodName{}, an adaptation of UniVTG designed to tackle \newTaskNameShort. 
\methodName{} achieves high negative rejection accuracy (avg. $98.4\%$) scores while retaining moment retrieval scores to within $3.87\%$ Recall@1. Dataset splits and code are available at \url{https://github.com/keflanagan/MomentofUntruth}
\end{abstract}

\section{Introduction}
\label{sec:intro}
With the ever-increasing amount of video data that is accessible to the public through streaming websites, the ability to quickly search through this data is attaining an increased importance. Not only is it necessary to search for relevant videos, it is also desirable to search through the videos themselves. The video moment retrieval task addresses this by searching for relevant moments within videos using text queries as input. Currently, video moment retrieval models focus only on producing accurate start and end times for text queries, under the assumption that the moment \textit{always exists} within the video. Video moment retrieval datasets are composed of video-sentence pairs which all have a direct correspondence through labelled start and end times. 
However, this raises the question: \\
\textit{How do models perform with irrelevant textual queries for a given video?}

\begin{figure}[t]
    \centering
    \includegraphics[width = \linewidth]{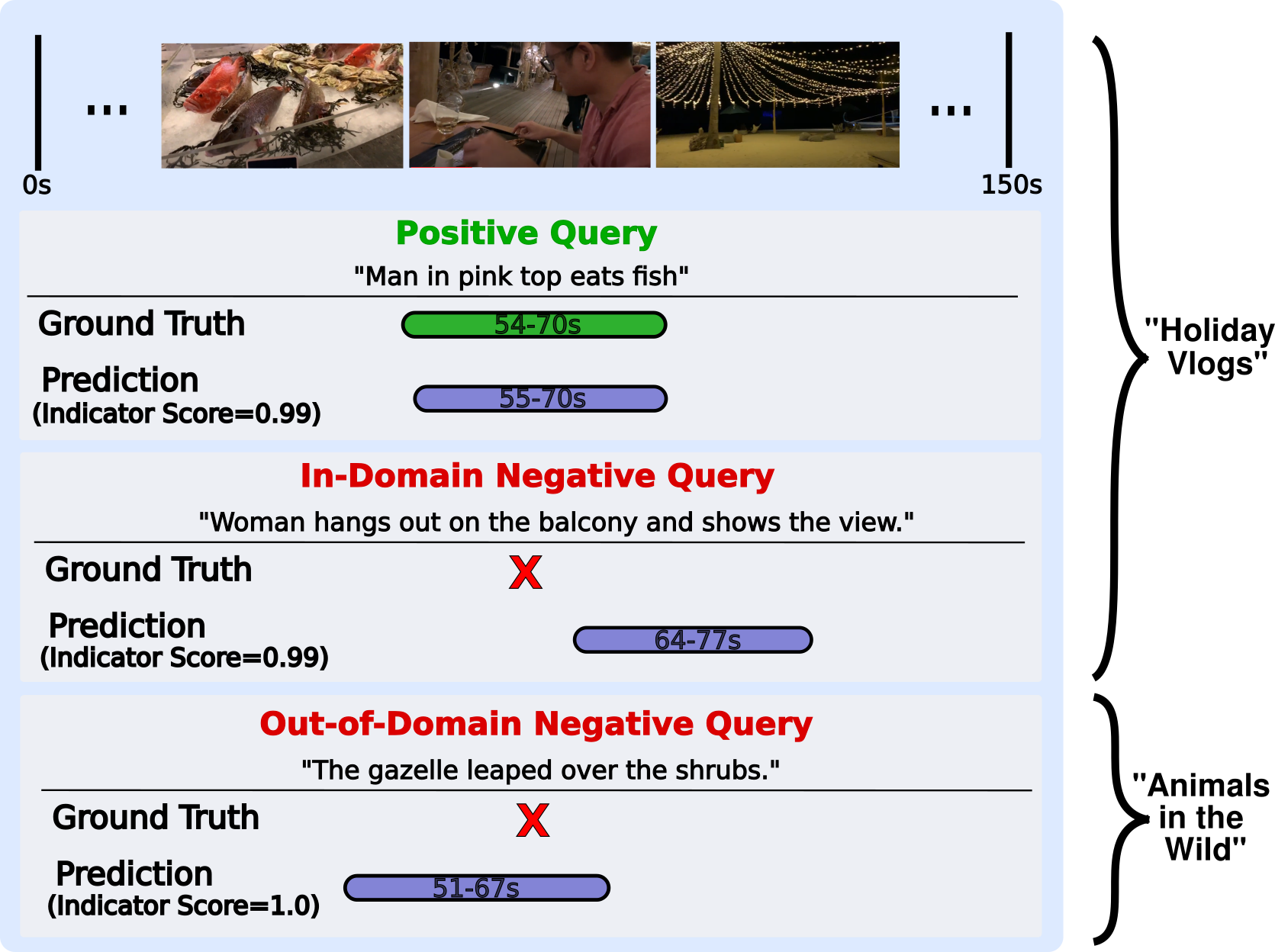}
    \caption{Video moment retrieval models are designed to predict start and end times in a video given a query sentence. Regardless of whether the text query is positive (exists in the video), in-domain negative (from the same domain but not present) or out-of-domain negative (from an entirely different scenario), current models such as UniVTG still produce a start-end time prediction.}
    \label{fig:intro}
        \vspace*{-12pt}
\end{figure}

For example, suppose you have a video of a person eating dinner and then you ask the model to localise the sentence \textit{``the gazelle leaped over the shrubs''}. Should a video moment retrieval model return a start and end time for this sentence even though it clearly doesn't correspond with anything in the video?
Current models will always provide a start and end time prediction regardless of the relevancy of the sentence, which can lead to hallucination (see Figure \ref{fig:intro}). Whilst there has been a great amount of progress in improving these models, they do not handle sentences which are irrelevant to the video, known as negative queries. We believe that if these models are to be robust and useful, they should be able to indicate when an input text query is irrelevant to the given video. We believe this task of negative rejection is a crucial aspect to ensure trustworthy and explainable AI models.

In this paper, we demonstrate, quantitatively and qualitatively, that current methods are not robust to negative queries and are not designed to distinguish between positive and negative queries.
We do this by testing a combination of both in-domain, i.e. queries from the same dataset not relevant to the query video, and out-of-domain negatives, i.e. queries from a different scenario.
We highlight examples of In-Domain (ID) and Out-Of-Domain (OOD) negatives within Figure~\ref{fig:intro} with example failure cases.

To combat this, we propose a method to reject negative queries while maintaining video moment retrieval performance. Our method uses an additional head which is explicitly trained to predict whether the sentence is relevant.
We train the model with both ID and OOD negatives, showing both are necessary.

Overall, our contributions are as follows: (i) To the best of our knowledge, we conduct the first analysis of the robustness of SOTA video moment retrieval models to negative queries, while making the important distinction between in-domain and out-of-domain negative queries. 
(ii)~We outline a method for sampling in-domain and out-of-domain negative queries for use during training and evaluation and provide new evaluation benchmarks for two popular Video Moment Retrieval datasets: QVHighlights~\cite{lei2021qvhl} and Charades-STA~\cite{gao2017tall}.
(iii) We propose a new framework which enables existing video moment retrieval methods to train for \newTaskName{}.
(iv) We showcase strong results on negative rejection while retaining high moment retrieval performance from a model which has been adapted under our framework.

\section{Related Work}
\label{sec:related_work}
\myparagraph{Moment Retrieval.} 
The task of video moment retrieval was first introduced in~\cite{gao2017tall, anne2017localizing}, with the aim of expanding the action localisation task~\cite{Onea2013ActionAE, shou2016temporal} to the open vocabulary setting. Moment retrieval methods require cross-domain interactions between video and text in order to determine the correspondences between them. Approaches have historically been divided largely between proposal-based~\cite{anne2017localizing,cao2021pursuit,gao2017tall,ge2019mac,jiang2019cross,liu2018cross,liu2021context,liu2022exploring,xu2019multilevel,chen2018temporally,qu2020fine,wang2020temporally,zhang20192dtan} and proposal-free methods~\cite{chen2020learning,chen2020hierarchical,chen2020rethinking,chen2021end,hao2022query,liu2022memory,lu2019debug,nan2021interventional,rodriguez2020proposal,yuan2019find,zhang2021natural,zhao2020bottom, zhang2020span}. 

Recent works have been designed to jointly perform moment retrieval and highlight detection~\cite{lei2021qvhl, han2024unleash}, being trained with both \textbf{moment start/end times} and \textbf{saliency scores} which are the ground truths for the highlight detection task. Certain datasets such as QVHighlights~\cite{lei2021qvhl} contain both human-annotated moments and saliency scores. Datasets which do not have human-annotated saliency scores instead use pseudo-saliency scores from the moment start/end times, with a non-zero score within ground truth moments.

\myparagraph{Moment Retrieval Methods.}
Moment-DETR~\cite{lei2021qvhl} is one such method that jointly trains on highlight detection and has served as a base for many recent moment retrieval methods~\cite{liu2022umt, moon2023query, moon2024cgdetr, lin2023univtg, jang2023knowing, lee2023bamdetr, xu2023mh, sun2024tr, liu2024r, xiao2024bridging}. 
This approach takes inspiration from DEtection TRansformer (DETR)~\cite{carion2020detr} methods in object detection, viewing moment retrieval as a direct set prediction problem, generating moment candidates and associated scores with which to rank them in an end-to-end manner. It concatenates text and video features as input and passes them through a transformer encoder-decoder. Trainable positional embeddings known as \textit{moment queries} are inputted to the decoder to generate candidate moment predictions. It has separate prediction heads for the moment span; the score of each proposed moment; and the saliency scores. While Moment-DETR uses foreground/background labels during training to supervise the moment score predictions, these are not designed to be used during evaluation for determining positive/negative queries.
Some approaches based on this framework have included the addition of extra prior information to initialise the queries~\cite{jang2023knowing} or the addition of an extra modality~\cite{liu2022umt}. Others have focused on improving the video-text interactions in the model~\cite{moon2023query, moon2024cgdetr}.

UniVTG~\cite{lin2023univtg} jointly trains for three tasks: moment retrieval, highlight detection, and video summarisation from datasets which do not contain a training signal for all three. It utilises a large pre-training scheme and alters the Moment-DETR architecture by removing the decoder and estimating an indicator score and predicted span for every feature clip in the video. This replaces the moment queries used within Moment-DETR.  
Furthermore, saliency scores are produced purely through feed-forward layers and attentive pooling over text features, without being passed through the transformer encoder.

QD-DETR~\cite{moon2023query} alters the encoder to contain cross-attention layers to ensure that the visual features are being properly attended to by the text features. It makes use of negative queries during training, shuffling video-sentence pairs across the dataset. This forces the saliency score predictions to be more indicative of the relevance of the sentence to the video clips.
However, saliency score predictions are still not used for the video moment retrieval task.

CG-DETR~\cite{moon2024cgdetr} employs a clip-word correlation learner and uses dummy tokens concatenated with the text query tokens. These dummy tokens are designed to attend to sections of the video that are not represented by the query, thus essentially representing the query-excluded meaning. This helps to prevent irrelevant video clips from being represented by the text query.

Despite progress in moment retrieval performance, \emph{all methods assume queries at inference are always positive}. 
Models are not evaluated with an input text query that is irrelevant to the video. 
Additionally, models are not equipped to handle such negative queries.

\myparagraph{Video Corpus Moment Retrieval.}
Video Corpus Moment Retrieval (VCMR)~\cite{escorcia2019eemporallo, lei2020tvr, zhang2021video, hou2021conquer} is another related task. This expands the search from a single video to a corpus of videos. While this returns only a single video segment from a set of videos, essentially serving as a rejection of the other videos, VCMR works under the assumption that the query is present in exactly one video in the corpus. There is no active scheme to determine negatives. Currently, VCMR methods cannot determine if a query is \emph{not} present in any videos in the corpus, similar to the current state of video moment retrieval. Therefore this task formulation also does not allow for negative rejection in video moment retrieval.

\myparagraph{Negative Rejection in Other Tasks.} 
Modelling uncertainty of predictions, and discarding decisions with low certainty has been integrated into classification tasks~\cite{pudil1992reject, radu2006reject, wiener2011agnosticsc, franc2023optimal}.
Our approach differs from these because the model is actively predicting whether the query is present in the video, rather than stating that it is unsure about the prediction.

Recently, the topic of negative rejection has been highlighted~\cite{chen2024benchmarking} in LLMs with Retrieval Augmented Generation, whereby information is extracted from retrieved documents to aid in responding to input queries. Negative rejection is crucial in cases where the required information is not present in the retrieved documents, as otherwise the hallucination of incorrect information can occur. It has been shown that currently these LLMs are not robust to negative rejection~\cite{chen2024benchmarking}. Our findings for the video moment retrieval task parallels this study, where current models are hallucinating moments in videos when negative queries are provided. Just as it will be important to deal with this issue in LLMs in order to improve reliability, it will also be important to achieve negative rejection in video moment retrieval.

\section{Method}
\label{sec:method}

In this Section, we first present details of the standard Video Moment Retrieval Task and how \newTaskName{} differs in Sec.~\ref{subsec:vmr_task}, before defining types of negative queries in Sec.~\ref{subsec:neg_queries} and how they can be collected in Sec.~\ref{subsec:create_neg_query}. Lastly, we detail how current Video Moment Retrieval models can be trained for \newTaskName{} in Sec.~\ref{subsec:vmr_method}.

\subsection{\newTaskName}
\label{subsec:vmr_task}
We first present the Video Moment Retrieval task as defined in the literature.
Formally, for each video $V_i$ within a corpus, there exists a set of query sentences $q_{i,j}$ with corresponding moments given as start $t^s_{i,j}$ and end $t^e_{i,j}$ times.
We collectively describe this as the set of queries for video $V_i$: $Q_i=\{(q_{i,j}, t^s_{i,j}, t^e_{i,j})\}$.
During training, models learn to predict the start/end times of a moment given the corresponding query sentence for the $i$th video.

At inference time, methods are evaluated on their ability to correctly localise the query sentence $q_{i,j}$ which is always assumed to be contained within video $V_i$.
Therefore, methods rank and select the highest proposal/predicted moment $(\Tilde{t}^s_{i,j}, \Tilde{t}^e_{i,j})$ and compare this to the ground truth moment $(t^s_{i,j}, t^e_{i,j})$ directly during evaluation.
By doing so, all Video Moment Retrieval methods make the assumption that all query sentences $q_{i,j}$, positive or negative, are relevant and contained within a video $V_i$.

In this work, we propose \newTaskName{} (\newTaskNameShort) in which models should reject negative query sentences which are not contained within the video and return a moment span only for positive query sentences that are contained within the video. 
The model therefore predicts the start/end times as before as well as whether to accept or reject the query.
Formally, the model will output a tuple $(\tilde{y}, \tilde{t^s}, \tilde{t^e})$ containing the prediction score, $\tilde{y}$, and the predicted start/end times $\tilde{t}^s$ and $\tilde{t}^e$.
In the case of a negative prediction $\tilde{y}=0$, then $\tilde{t}^s$ and $\tilde{t}^e$ are considered invalid and rejected.
For a positive prediction score ($\tilde{y}=1$) the predicted start/end times are considered valid and compared to the ground truth moments as normal.

For both training and evaluation both positive and negative queries need to be utilised.
We define positive queries for video $V_i$ as a tuple of $Q^+_i=\{(y_{i,j}=1, q_{i,j}, t^s_{i,j}, t^e_{i,j})\}$, and negative queries $Q^-_i=\{(y_{i,k}=0, q_{i,k})\}$.
Next, we provide more detail regarding the negative queries.

\subsection{In-Domain \& Out-of-Domain Negatives}
\label{subsec:neg_queries}
We choose to divide negative queries into two categories, namely \textbf{In-Domain (ID)} and \textbf{Out-of-Domain (OOD)}, which represent queries from a similar context to the video and queries from a different context. 
The distinction is related to the plausibility of the query being present in the video. 
Both sets of negatives are important to consider when examining the behaviour of moment retrieval models. ID negatives allow for the inspection of a model's ability to differentiate specific details in videos, while OOD negatives enable the inspection of a model's ability to recognise that a query is entirely irrelevant to the scenario.
Regardless of whether the negative is in-domain or out-of-domain, a \newTaskName{} method should correctly recognise that there is no corresponding moment to retrieve from the video.

\textbf{In-Domain Negatives} are queries describing events which do not occur in a given video, but which feasibly could occur within a video from that domain or context. For example, in videos of a person carrying out actions in a kitchen, the sentence ``the person opens the oven'' would be plausible, and would be an ID negative if no oven was opened within this video.

\textbf{Out-of-Domain Negatives} are defined as queries which belong to an entirely different scenario to the selected video, and which are therefore extremely unlikely to be present within it. An example is the sentence ``the player hits the ball across the net'' for the above video of a person carrying out actions in a kitchen.

\subsection{Sampling Negative Queries}
\label{subsec:create_neg_query}
\begin{figure}[t]
    \centering
    \includegraphics[width = \linewidth]{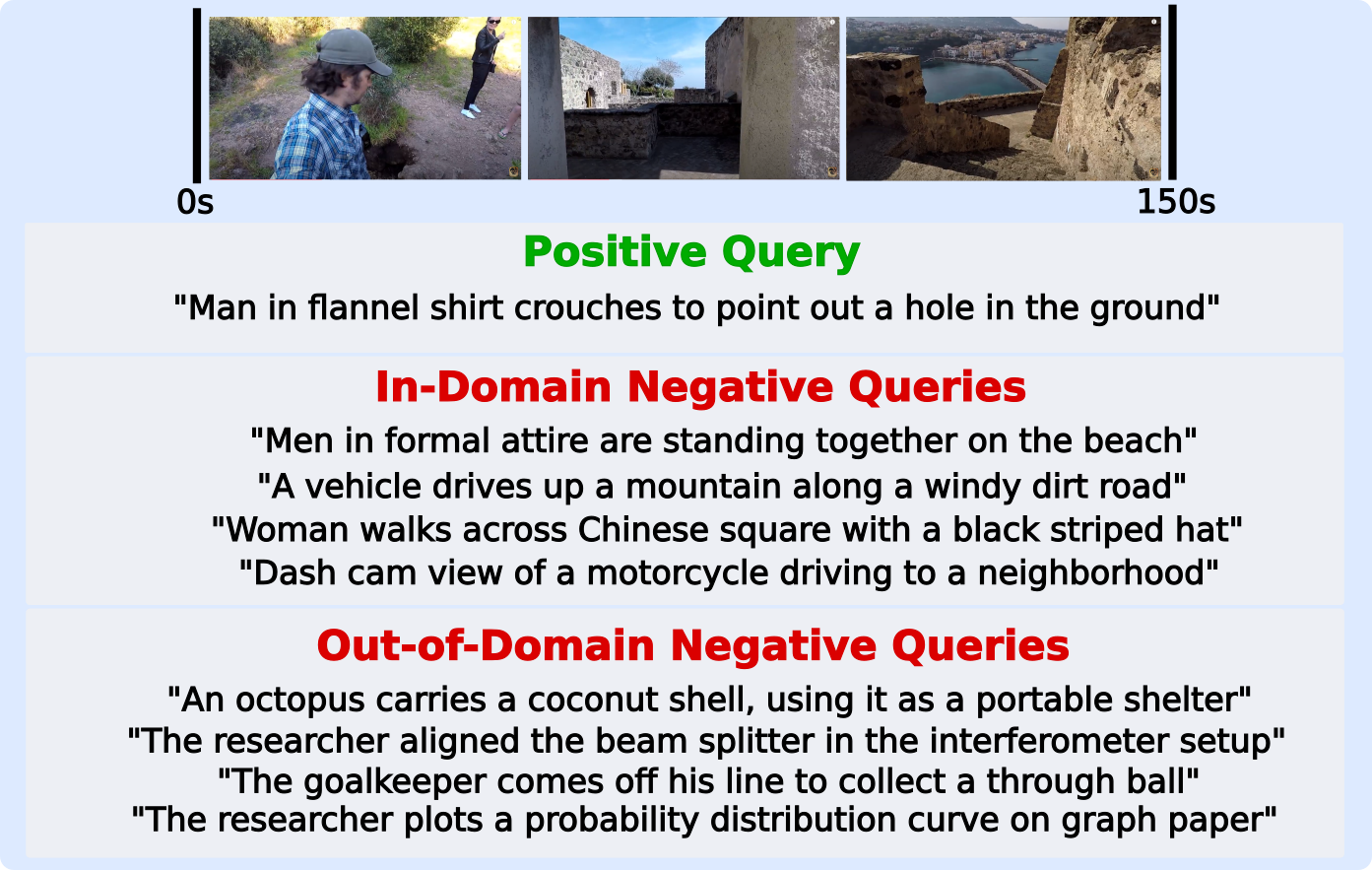}
    \caption{In-Domain and Out-of-Domain negative queries alongside a corresponding positive video-sentence pair.}
    \label{fig:negative_examples}
        \vspace*{-12pt}
\end{figure}

We sample ID and OOD negatives to ensure that these represent true negatives whilst still including a wide variety of queries. Examples of these are displayed in Figure \ref{fig:negative_examples}.

\myparagraph{In-Domain Negatives} are produced by shuffling video-sentence pairs within the dataset, similar to the method used in~\cite{moon2023query}, assigning a different video to each sentence.
We choose not to generate new ID negatives to reduce the chance of false negatives in the test set, which occur when a moment described by that query sentence is present in the newly assigned video.
We ensure a clean test set as follows:
Firstly, cosine similarity scores are calculated for each sentence with every other sentence in the test set by using CLIP embeddings.
Then, a pseudo-similarity score is calculated between a sentence and every video in the test set by determining the highest sentence-sentence similarity score for each video.
For example, to get the similarity score between query sentence $q_k$ and video $V_i$, you first calculate the cosine similarity between $q_k \notin Q_i$ and each $q_{i,j} \in Q_i$ and take the maximum of those similarity scores to be the sentence-video pseudo-similarity score.
Each sentence is then assigned to a video whose video-sentence similarity score is in the lowest 50th percentile for that sentence.

\myparagraph{Out-of-Domain Negatives} are generated via a large language model (LLM).
Scenarios are selected whose actions would be extremely unlikely to occur within the specific dataset.
Within these scenarios, the LLM is prompted to generate sentences describing actions from a variety of subtopics within the scenario.
For example, for a cooking dataset, the scenario of competitive sport might be chosen.
Accordingly, the LLMs are prompted to generate sentences describing actions occurring in topics such as ``football", ``basketball'', ``tennis'' and others. 
We select the scenarios based on the dataset to ensure that there is no overlap and generated sentences are not false negatives.

\subsection{Modelling with Negative Queries}
\label{subsec:vmr_method}
Many current Video Moment Retrieval methods are based on Moment-DETR~\cite{lei2021qvhl}.
We detail the general concept of these methods first before describing extensions towards \newTaskName.

\myparagraph{Video Moment Retrieval Methods.}
Current Moment-DETR-based methods utilise frozen video and text encoders, usually followed by a projection layer to match dimensionality.
This produces video features $\textbf{V} = \{\textbf{v}_c\}_{c=1}^{L_v}$ and text features $\textbf{Q} = \{\textbf{q}_w\}_{w=1}^{L_q}$ where 
$L_v$ is the number of video clips and $L_q$ is the number of query sentence tokens.
Methods typically pass these video and text features into a Transformer encoder to produce text-attended video tokens. It is common to use a Transformer decoder with $M$ trainable position embeddings known as \textit{moment queries} as input alongside the text-attended video tokens, thus producing $M$ final video representations which may be used as input to the heads. Certain methods such as UniVTG~\cite{lin2023univtg} instead directly utilise the text-attended video tokens as input to the three predictions heads. 

These methods use three prediction heads. Firstly the foreground matching head produces the indicator scores $\tilde{f}_m$, where $m$ is the index of the candidate moment predictions. These are typically produced by a set of feed forward layers and an activation function on top of the $M$ video representations and aim to predict the likelihood of the moment matching the query.
Similarly, the moment boundaries $\tilde{d}_m$ are predicted via another boundary prediction head, but with two outputs: either start/end time offsets or moment centre and width. 
There can also be a saliency head to predict the saliency scores $\tilde{s}_c$, where $c$ is the clip index. The input to the saliency head is typically the output of the video encoder or earlier representations. The saliency scores are used only for the highlight detection task, which aims to detect text-guided highlights for a given video.

Whilst specific details of how these moment and saliency predictions are produced vary, the basic principles outlined above remain the same.

\begin{figure}[t]
    \centering
    \includegraphics[width = 0.95\linewidth]{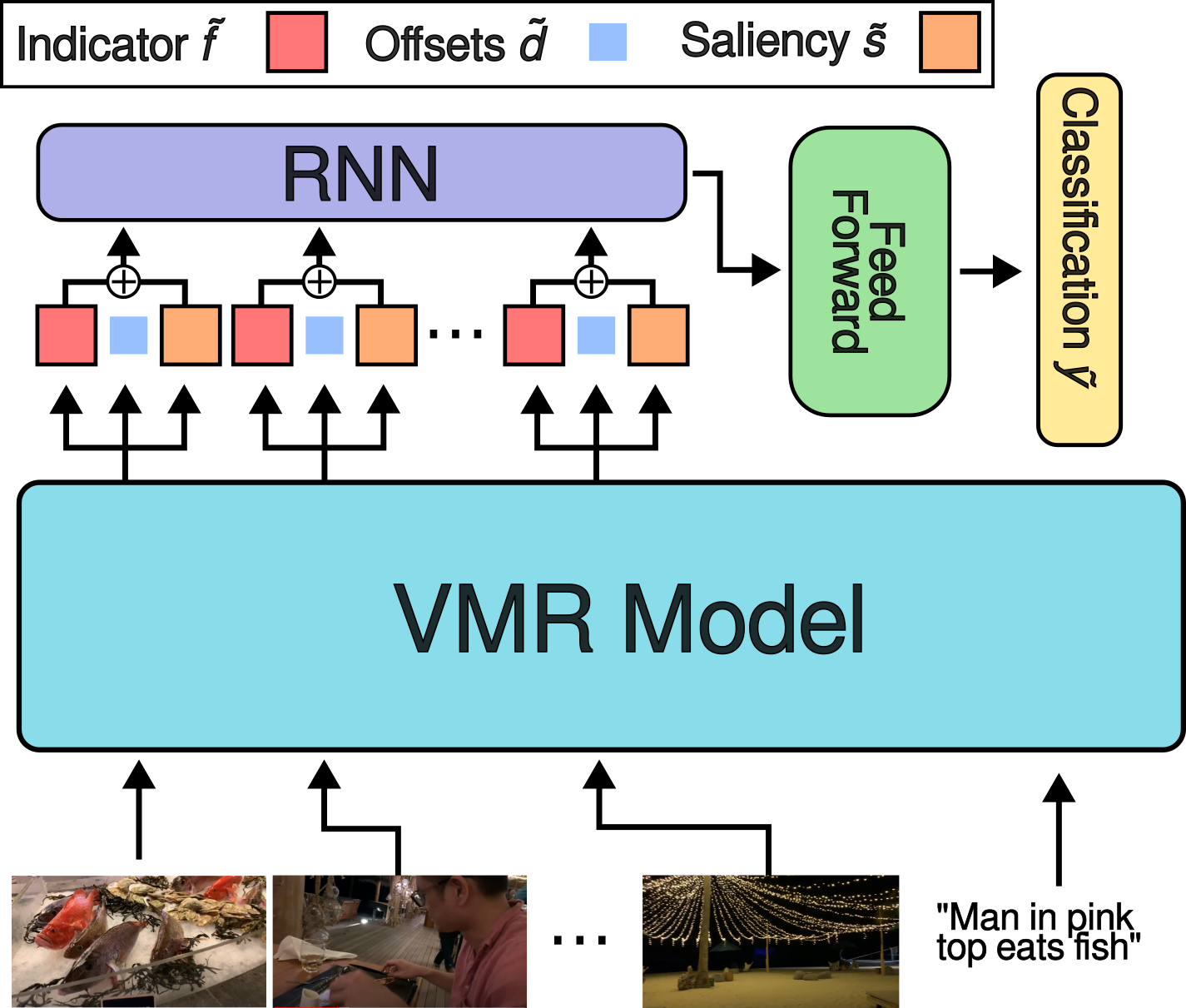}
    \caption{The classification head for \newTaskNameShort{} is added to a Video Moment Retrieval model (UniVTG in this case) via summation of the indicator and saliency scores, which are then passed through a recurrent layer and a feed forward layer before producing a single value output for classification.}
    \label{fig:method}
        \vspace*{-12pt}
\end{figure}

\begin{figure*}[t]
    \centering
    \includegraphics[width = \linewidth]{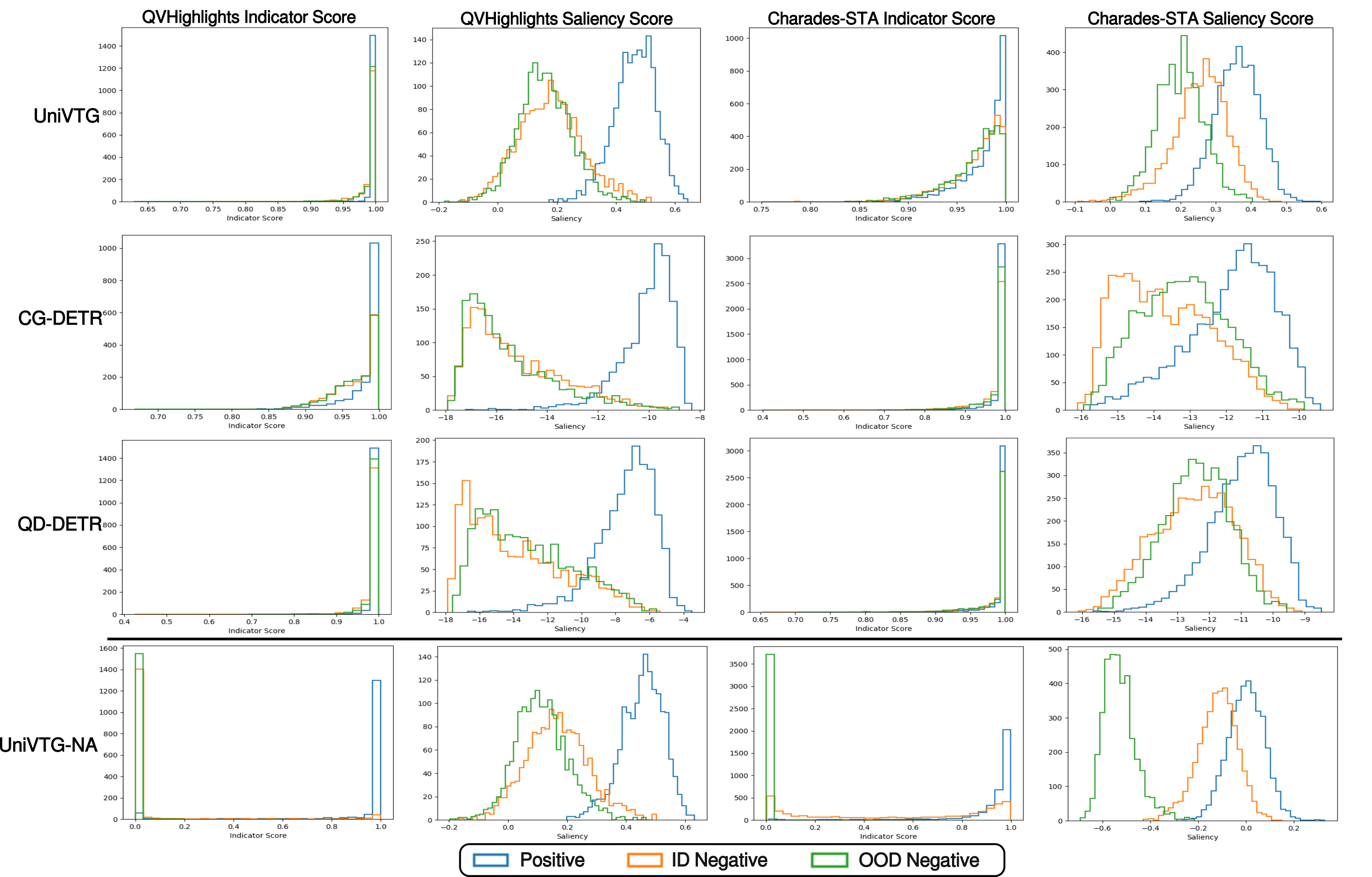}
    \caption{Histograms of indicator and saliency scores from UniVTG, CG-DETR and QD-DETR on the positive and negative queries. Bottom row: Indicator and saliency scores from \methodName{}.}
    \label{fig:model_histograms}
        \vspace*{-12pt}
\end{figure*}

\myparagraph{Incorporating Negative Queries.}
We propose to alter Moment-DETR-based methods as follows.
The generated negative queries are used to train the model to differentiate between positive and negative video-sentence pairs, enabling \newTaskName{}.
As shown in Figure~\ref{fig:method}, we add a binary classification head on top of the base Video Moment Retrieval model which classifies queries as positive or negative. The indicator score and saliency score predictions are combined and passed into a recurrent (RNN) layer. Both are used, as while the indicator scores denote the likelihood of the moment matching the query, saliency scores from Figure \ref{fig:model_histograms} are much more discriminative between positive and negative.
We employ an RNN to maintain temporal knowledge within the features while handling variable video lengths in a lightweight manner.
The RNN output at the final step is then passed into a feed-forward layer (MLP) and, finally, a sigmoid activation function to produce the final prediction score, $\tilde{y}$.
\begin{equation}
    \tilde{y} = \sigma(\text{MLP}(\text{RNN}(\tilde{f} \oplus \tilde{s})))
    \label{eq:class_head}
\end{equation}
where, the $\oplus$ operation in Equation \ref{eq:class_head} represents a generic combination. This can be a concatenation, learned combination, or in our case, a summation of the indicator and saliency scores. We denote models following this architecture \textit{Negative-Aware} with \textit{NA} as the suffix.

\myparagraph{Losses.}
A binary cross entropy loss is applied to the classification head output, with positive queries taking a ground truth value of $y = 1$ and negative queries taking a value of $y = 0$. $\tilde{y}$ is the output prediction from the classification head and $\lambda_p$ is the loss weighting. The loss is given as:
\begin{equation}
    \mathcal{L}_p = \lambda_p(y\log\tilde{y} + (1-y)\log(1-\tilde{y}))
    \label{eq:bce_loss}
\end{equation}

Typically, methods utilise a foreground matching head loss $\mathcal{L}_f$ on the indicator scores, a boundary loss $\mathcal{L}_b$ across the start/end times, and a saliency loss $\mathcal{L}_s$ applied to the saliency scores.
For positive queries, these losses can be applied as normal.
However, losses on negative queries are adapted as follows: $\mathcal{L}_b$ is set to 0 as there is no ground truth boundary to predict,
while all ground truth values for the indicator and saliency scores are set to 0 for $\mathcal{L}_f$ and $\mathcal{L}_s$.
These loss functions may need to be adjusted if for example they involve contrastive losses, and as such are denoted $\mathcal{L}_f^-$ and $\mathcal{L}_s^-$ for negatives. This follows the understanding of their basic function; to lower $\tilde{f}_i$ and $\tilde{s}_i$ predictions for negative queries.

The original losses are combined with the proposed classification loss $\mathcal{L}_p$, which defines the losses for positives $\mathcal{L}^+$ and negatives $\mathcal{L}^-$.
\begin{equation}
    \mathcal{L}^+ = \mathcal{L}_f + \mathcal{L}_b + \mathcal{L}_s + \mathcal{L}_p
\qquad
    \mathcal{L}^- = \mathcal{L}_f^- + \mathcal{L}_s^- + \mathcal{L}_p
    \label{eq:sep_losses}
\end{equation}
The total loss $\mathcal{L}_{tot}$ is the weighted sum of the losses for positives $\mathcal{L}^+$, ID negatives $\mathcal{L}_{ID}^-$, and OOD negatives $\mathcal{L}_{OOD}^-$, where $\lambda^+$, $\lambda_{ID}^-$ and $\lambda_{OOD}^-$ are the loss weights and $\mathcal{L}_{ID}^-$ and $\mathcal{L}_{OOD}^-$ are $\mathcal{L}^-$ applied to the specific domains. 
\begin{equation}
    \mathcal{L}_{tot} = \lambda^+\mathcal{L}^+ + \lambda^-_{ID}\mathcal{L}^-_{ID} + \lambda^-_{OOD}\mathcal{L}^-_{OOD}
    \label{eq:losses_pos_neg}
\end{equation}
While the exact losses may vary depending on the model architecture, we have described the basic principles of adjusting these losses to account for negative queries. 

To summarise, we add an extra classification head on top of the indicator and saliency score predictions. This head classifies the query sentence as positive or negative. During training, the saliency and indicator scores are set to 0 for the negative queries and the losses are adjusted where necessary to ensure that the model is able to learn a strong signal to differentiate between positive and negative queries. The details for each model are found in Sec.~\ref{sec:model_detail} of the Appendix.

\section{Experiments}
We first present information on the evaluation protocol, giving details of the metrics and datasets used; implementation details; and baseline implementations.

\myparagraph{Metrics.}
We use the standard moment retrieval metric of Recall@k with IoU@$\theta$, following the literature~\cite{lei2021qvhl, lin2023univtg, moon2023query, moon2024cgdetr},  in specifying k=1 and $\theta\in\{0.5, 0.7\}$.  We also introduce the metric of \textbf{Rejection Accuracy (RA)} which is the percentage of negative queries correctly rejected. 

\myparagraph{Models.} We report results on UniVTG~\cite{lin2023univtg},  CG-DETR~\cite{moon2024cgdetr}, and QD-DETR~\cite{moon2023query}, alongside their Negative-Aware adaptations, focusing primarily on \methodName{}.

\begin{table*}[t]
\begin{minipage}{0.97\textwidth}
    \centering
    \resizebox{\columnwidth}{!}{
    \begin{tabular}{lccccrrlccrr}
    \toprule
           &            &            &\multicolumn{4}{c}{QVHighlights}&&\multicolumn{4}{c}{Charades-STA}\\ \cmidrule{4-7} \cmidrule{9-12}
           &            &            &     &     &\multicolumn{2}{c}{Rejection Acc. (\%)}&&     &     &\multicolumn{2}{c}{Rejection Acc. (\%)}\\ \cmidrule{6-7} \cmidrule{11-12}
    Method &$\tilde{f}$ &$\tilde{s}$ &R1@0.5&R1@0.7&\multicolumn{1}{c}{ID}&\multicolumn{1}{c}{OOD}&&R1@0.5&R1@0.7&\multicolumn{1}{c}{ID}&\multicolumn{1}{c}{OOD}\\ \midrule
    UniVTG-Thr~\cite{lin2023univtg} &$\checkmark$&$\times$    &67.23 (\textcolor{red}{-0.12}) &52.52 (\textcolor{red}{-0.13})&  8.97             &  6.52          &&60.19 (\textcolor{red}{-0.03})&38.55 (\textcolor{green}{+0.00})&  0.56             &  0.24          \\
    CG-DETR-Thr~\cite{moon2024cgdetr} &$\checkmark$&$\times$    &  67.03 (\textcolor{red}{-0.07})   &  53.55 (\textcolor{green}{+0.00})   &      0.32             &      0.39          &&     57.02 (\textcolor{red}{-0.51}) &  35.43 (\textcolor{red}{-0.24})   &        23.15           &   9.52             \\
    QD-DETR-Thr~\cite{moon2023query} &$\checkmark$&$\times$    &  61.94 (\textcolor{red}{-0.06}) & 46.19 (\textcolor{red}{-0.07})  &  0.39   &       0.06                     &&  58.92 (\textcolor{red}{-0.19})   &  36.64 (\textcolor{red}{-0.11})   &       4.19         &       4.57       \\ \midrule
    UniVTG-Thr~\cite{lin2023univtg} &$\times$    &$\checkmark$&67.35 (\textcolor{green}{+0.00})&52.65 (\textcolor{green}{+0.00})& 64.65      & 73.03          && 60.05 (\textcolor{red}{-0.17})& 38.47 (\textcolor{red}{-0.08})&9.81             &19.95          \\
    CG-DETR-Thr~\cite{moon2024cgdetr}&$\times$    &$\checkmark$&  66.58 (\textcolor{red}{-0.52})   &  53.23 (\textcolor{red}{-0.32})   &     82.00              &     86.45           &&    56.85 (\textcolor{red}{-0.68}) & 35.43 (\textcolor{red}{-0.24})    &          10.86         &    6.29            \\
    QD-DETR-Thr~\cite{moon2023query}&$\times$    &$\checkmark$&  57.23 (\textcolor{red}{-4.77})   &  43.61 (\textcolor{red}{-2.65})   &  89.68                 &    87.35            &&    58.39 (\textcolor{red}{-0.72}) &  36.34 (\textcolor{red}{-0.41})   &        15.51        &      23.04          \\
    \bottomrule
    \end{tabular}}
    \caption{Video Moment Retrieval results using a threshold on the indicator score ($\tilde{f}$) and saliency score ($\tilde{s}$). Thresholds are chosen by setting the 0.5th percentile on the positive training set. The differences to R1@$\theta$ with no negative rejection are highlighted alongside.}
    \label{tab:no_training_results}
\end{minipage}\hfill
\vspace*{-12pt}
\end{table*}

\myparagraph{SVM Baseline.}
We use SVMs as an alternative classifier for negative rejection.
Saliency score outputs from the original model trained without negative query classification are passed through a Support Vector Machine (SVM).
We use both ID and OOD negatives to train this SVM.
For each video and query sentence pair in the training set, the top 3 predicted saliency score values across the video are found and averaged to provide a score of relatedness between query sentence and video.
These values are passed into the SVM to train it to classify each query as either positive or negative.
The trained model is then applied to test data to produce positive/negative classifications.

\myparagraph{Datasets.}
We use the QVHighlights \cite{lei2021qvhl} and Charades-STA \cite{gao2017tall} datasets, following \cite{lei2021qvhl}. QVHighlights consists of vlogs and news report videos sourced online, providing both human annotated moment start/end times and saliency scores for each video-sentence pair. We report results on the publicly available validation set of QVHighlights. Charades-STA is made up of home videos with scripted actions and provides just moment start/end times.

\myparagraph{Negative Queries.}
Our OOD train and test sets have 7230 and 1550 text queries respectively, to match the numbers from the QVHighlights train and val sets. For the larger Charades-STA, the batch sizes are retained during training so some OOD negative queries are sampled twice during each epoch, but are assigned to random videos so produce a distinct training signal during each iteration. For the test set, OOD negative queries are assigned to multiple videos in order to match the larger number of positive queries in Charades-STA. ID negative sets always match the size of the positive sets. 

\myparagraph{Implementation Details.}
We use the same pre-trained model provided with UniVTG \cite{lin2023univtg} as the base model for training UniVTG-NA on each dataset.
For each model, RNN and feedforward layers all have a hidden dimension of 50.
Batch sizes of 32 each are used for the positive, ID negative, and OOD negative queries during training. Loss weight values are found in Sec.~\ref{subsec:implementation_details} of the Appendix.

For OOD negative query generation, the broad topics of ``animal behaviour'', ``competitive sports'', ``physics laboratory'' and ``mathematics class'' are used. Within each scenario, more specific subtopics are used to generate queries. The subtopics are selected so as to produce a broad range of sentences across the scenarios. The LLMs used for generation are Claude-3-Opus and GPT-4o. Details of the prompts used and subtopics are found in Sec.~\ref{subsec:neg_query_generation} of the Appendix.

\begin{table*}[ht]
    \centering
    \begin{tabular}{lccrrlccrr}
    \toprule
                &\multicolumn{4}{c}{QVHighlights}&&\multicolumn{4}{c}{Charades-STA}\\ \cmidrule{2-5} \cmidrule{7-10}
                &     &     &\multicolumn{2}{c}{Rejection Acc. (\%)}&&     &     &\multicolumn{2}{c}{Rejection Acc. (\%)}\\ \cmidrule{4-5} \cmidrule{9-10}
    Method      &R1@0.5&R1@0.7&\multicolumn{1}{c}{ID}&\multicolumn{1}{c}{OOD}&&R1@0.5&R1@0.7&\multicolumn{1}{c}{ID}&\multicolumn{1}{c}{OOD}\\ \midrule
    \rowcolor{lightgray}
    UniVTG~\cite{lin2023univtg}      &67.35&52.65&   0.00            &   0.00         &&60.22&38.55&  0.00             &  0.00          \\
    UniVTG-SVM  &63.48&49.87&  94.77            &  97.74         &&53.47&33.49& 35.89             & 50.40          \\
    \methodName &63.48&50.00&  96.84            & 100.00         &&55.54&35.11& 64.11             &100.00          \\
    \midrule
    \rowcolor{lightgray}
    CG-DETR~\cite{moon2023query}      &67.10&53.55&   0.00            &   0.00         &&57.53&35.67&  0.00             &  0.00          \\
    CG-DETR SVM &62.84 & 50.26 & 95.55 & 95.41         && 43.79& 28.44 & 82.9             & 74.52        \\
    CG-DETR-NA &63.03&49.42&  91.61& 99.94         && 50.05& 32.34 & 71.96             & 100.00          \\
    \midrule
    \rowcolor{lightgray}
    QD-DETR~\cite{moon2023query}      &62.00&46.26&   0.00            &   0.00         &&59.11&36.75&  0.00             &  0.00          \\
    QD-DETR SVM &48.26 & 37.42 & 96.32 & 95.09         &&46.67 & 30.05 & 76.91 & 81.59          \\
    QD-DETR-NA &59.10&44.52&  90.58            & 99.74         &&55.19&34.89& 55.91             &100.00          \\
    \bottomrule
    \end{tabular}
    \caption{Results of training using negative queries for the proposed \newTaskName{} task. We compare using an SVM on top of UniVTG~\cite{lin2023univtg} with the proposed method \methodName{} across both QVHighlights and Charades-STA. Our proposed method loses less R1@$\theta$ performance compared to using an SVM whilst improving upon rejection accuracy for both in-domain (ID) and out-of-domain (OOD) negatives. We also display corresponding results for CG-DETR-NA and QD-DETR-NA.}
    \label{tab:negative_training}
    \vspace*{-12pt}
\end{table*}

\subsection{Shortcomings of Current Methods}
\label{subsec:shortcomings}
In this section, we first explore how current methods that \emph{have not been trained} to explicitly distinguish negative queries are able perform at inference time on \newTaskName.
Specifically, we investigate current state of the art methods UniVTG~\cite{lin2023univtg}, CG-DETR~\cite{moon2024cgdetr}, and QD-DETR~\cite{moon2023query} on both QVHighlights~\cite{lei2021qvhl} and Charades-STA~\cite{gao2017tall}.

Table~\ref{tab:no_training_results} showcases an unsupervised approach, denoted with a suffix -Thr, for each of the three methods using either indicator scores ($\tilde{f}$) or saliency scores ($\tilde{s}$) as a threshold for determining positive or negative queries.
In both cases, thresholds were set to the lowest 0.5th percentile value of $\tilde{f}$ or $\tilde{s}$ in the training set of positive queries. This shows off-the-shelf performance from raw model outputs without requiring further modelling of the relationship between positive and negative query scores. The 0.5th percentile was chosen to set the classification boundary as the lower limit of the positive sentence scores, while mitigating the effect of outliers.
The results show that methods are unable to distinguish between positives and negatives when the indicator score is used.
Using saliency scores fares better for QVHighlights, though methods struggle on Charades-STA.

We analyse this further in Figure~\ref{fig:model_histograms} which shows histograms of the methods across both datasets looking at the indicator score and the saliency score of positive and negative queries within the test sets.
We note that the indicator score, which is the only score used for Video Moment Retrieval evaluation, is not separable and \emph{methods treat positives and negatives the same way}.
The saliency score provides more promise for separating positive and negative queries, but across all three methods there is still a considerable overlap between positives and negatives, especially on Charades-STA which doesn't have ground truth saliency scores to train on, leading to poor rejection accuracy.

\subsection{Training with Negative Queries}
\label{subsec:training_neg}
In this section, we utilise our proposed negative queries at training time. 
In Table~\ref{tab:negative_training} we compare using an SVM to predict whether a query is positive or negative vs. our negative-aware (NA) methodology.

We find that each negative-aware method achieves strong Rejection Accuracy performance while retaining much of the Recall@1,IoU@k performance. Across both QVHighlights and Charades-STA the OOD Rejection Accuracy for each method is $\sim$100\%, higher than the SVM. The ID Rejection Accuracy of each is over 90\% for QVHighlights. The lower scores on Charades-STA likely stem from similar actions occurring in many videos within Charades-STA. 
While the SVM achieves higher ID Rejection Accuracy on CG-DETR and QD-DETR, the Recall scores are generally much lower than the negative-aware methods. The reduction in Recall scores is largely due to false negative classification, where positive queries are incorrectly classed as negative and therefore are not included in the Recall score. The negative-aware methods retain moment retrieval performance close to the original model, while the SVM methods typically suffer a much larger decrease. This is particularly apparent for Charades-STA, where the lack of ground truth saliency scores makes it more difficult for the base models to learn to distinguish between positive and negative queries. The results show the particular importance of training with a classification head for datasets without
ground truth saliency scores available.
UniVTG-NA in particular improves on or matches the SVM across all metrics for both datasets, and shows the strongest overall performance in the combined moment retrieval and negative rejection tasks.

Overall, the performance on these datasets demonstrates the effectiveness of negative-aware training. Much of the video moment retrieval performance is retained while a large proportion of negative queries are rejected. While there is a trade-off between moment recall scores and negative rejection, this allows the model to be more robust to a wider variety of input queries and give more informative responses to those using the model. 
We now focus on~\methodName{} for qualitative results and ablation.

\begin{figure}[t]
    \centering
    \includegraphics[width=0.95\linewidth]{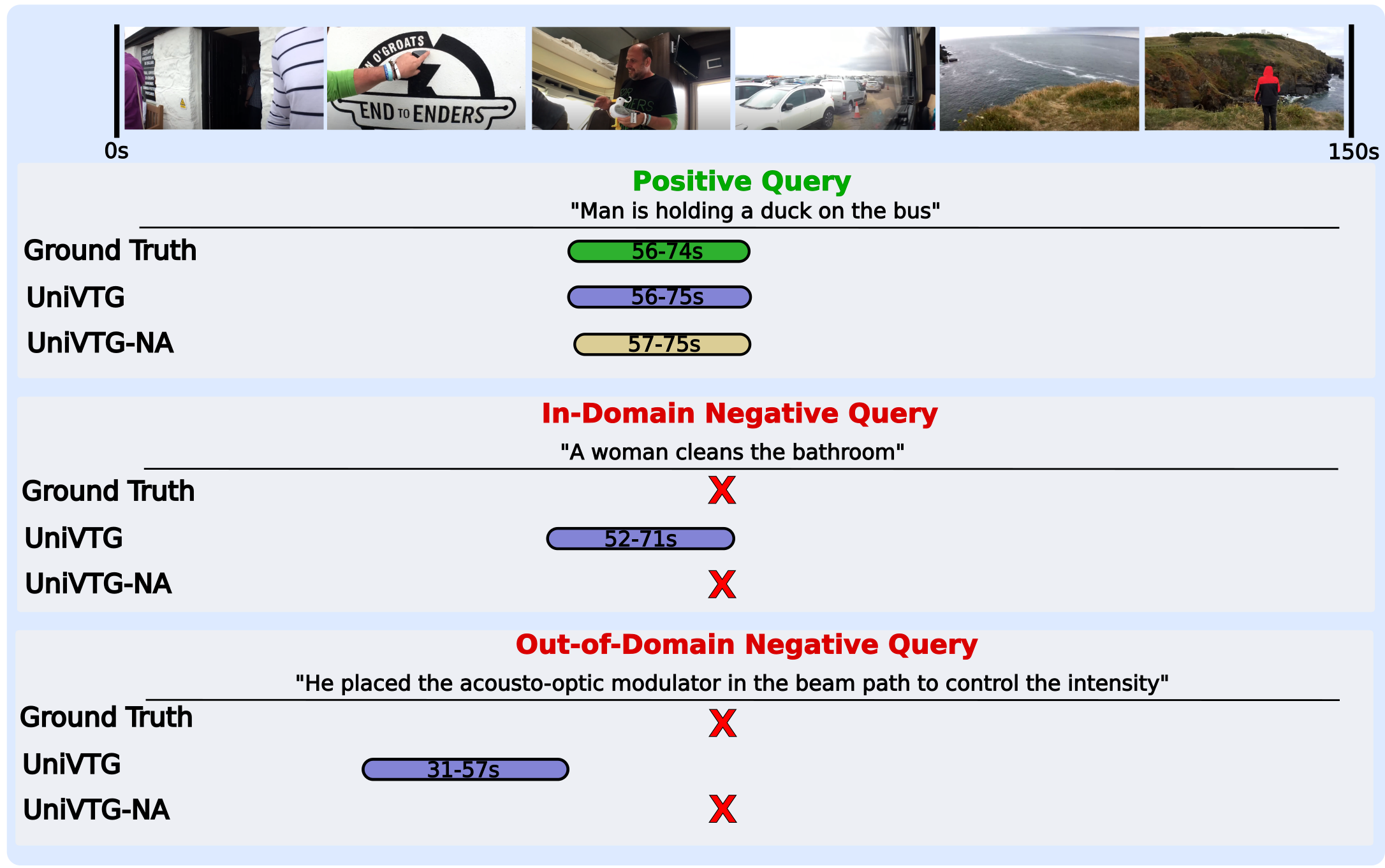}
    \caption{Qualitative result of \methodName{} on QVHighlights.}
    \label{fig:qual}
    \vspace*{-12pt}
\end{figure}

\myparagraph{Qualitative Results.}
Examples of predictions comparing \methodName{} and UniVTG are displayed in Figure \ref{fig:qual}. We see that \methodName{} retains accurate moment predictions while rejecting both ID and OOD negative queries, which UniVTG is not designed to do. Further qualitative results are found in Sec.~\ref{sec:qual} of the Appendix.

\myparagraph{Indicator and Saliency Scores for \methodName{}.}
We inspect the indicator and saliency score outputs of the trained \methodName{} model in  the bottom row of Figure \ref{fig:model_histograms}, to compare with the equivalent scores from the video moment retrieval models shown in the upper rows. We can see that through our training scheme, the indicator scores become strong signifiers of whether the query is positive or negative, with the scores becoming clearly separable. The saliency scores are also strong indicators, with the Charades-STA saliency scores being much more separable, enabling a clearer distinction between positive and negative queries despite the lack of ground truth saliency scores. This shows that our added classification head and adjusted negative loss training scheme 
allows the classifier head to learn to classify positive/negative queries while also enabling the indicator and saliency heads to learn and reinforce this signal. 

\begin{table}[t]
\centering
\resizebox{\columnwidth}{!}{%
\begin{tabular}{cc|cccc}
\hline
   \multicolumn{2}{c|}{\methodName{}} & R1@0.5 & R1@0.7  & \multicolumn{2}{c}{Rejection Accuracy (\%)}                      \\  & \\[-2.1ex] \cline{5-6}
  In-domain & Out-of-domain & & & In-Domain & Out-of-Domain\\ \hline
               $\checkmark$ &  $\checkmark$ & 63.48 & 50.0 & 96.84 & 100   \\ 
               $\checkmark$ &  - & 61.03 & 47.87 & 96.80 & 98.0 \\
               - &  $\checkmark$ & 66.45 & 52.71 & 32.8 & 100 \\ \hline
\end{tabular}}
\caption{Effect of including in-domain negatives and out-of-domain negatives in training on the QVHighlights dataset.}
\vspace*{-12pt}
\label{tab:neg_ablation}
\end{table}

\paragraph{Importance of In-Domain and Out-of-Domain.}
We present results in Table~\ref{tab:neg_ablation} of model performance when trained with just ID negatives or just OOD negatives. 
The model trained with just ID negatives achieves weaker performance on rejecting both ID and OOD negatives, while R1@k scores also decrease due to an increased number of false negative predictions. When trained with just OOD negatives, the model is able to retain its $100\%$ rejection of OOD negatives, but performance on ID negative rejection is much lower at $\sim33\%$. The increased R1@k score is a by-product of the weakened rejection ability, as fewer false negative predictions are made. The benefit of fewer false negatives is far outweighed by the detriment of low negative rejection.
These results indicate that ID and OOD negative queries offer a complementary training signal, mutually boosting performance. 

\myparagraph{Limitations.}
We note two limitations of the Negative-Aware approach: firstly, there remains a trade-off between localising positive moments and rejecting negative query sentences. Secondly, the pseudo-saliency scores used for datasets without ground truth human-annotated saliency scores are not as informative, which results in weaker negative rejection performance particularly in the in-domain case, as shown with the Charades-STA results.

\section{Conclusion}
In this work we have proposed the task of \newTaskName{}, which incoporates negative query rejection into the standard video moment retrieval task. We have analysed the ability of current video moment retrieval methods to adapt to \newTaskName{} and proposed negative-aware training which is specifically designed for this task. We have presented results for two new evaluation benchmarks on the QVHighlights and Charades-STA datasets. We have demonstrated the effectiveness of our method at rejecting negatives while maintaining high moment retrieval performance. 

\paragraph{Acknowledgments.} K Flanagan is supported by UKRI (Grant ref EP/S022937/1) CDT in Interactive AI \& Qinetiq Ltd via studentship CON11954. D Damen is supported by EPSRC Fellowship UMPIRE (EP/T004991/1) \& EPSRC Program Grant Visual AI (EP/T028572/1).

{\small
\bibliographystyle{ieee_fullname}
\bibliography{egbib}
}

\clearpage

\appendix

\section*{Appendix}
We provide more information about the dataset creation in Sec.~\ref{sec:dataset}, describing the process of generating the out-of-domain negative queries using LLMs and demonstrating our prompts and categories as well as details about the Negative-Aware Video Moment Retrieval dataset. We display full results for SVMs trained on output saliency scores for all three models (UniVTG~\cite{lin2023univtg}, CG-DETR~\cite{moon2024cgdetr}, QD-DETR~\cite{moon2023query}) in Sec.~\ref{sec:svms}. We expand on the adjustments made to the losses for UniVTG-NA and detail the QD-DETR and CG-DETR implementations in Sec.~\ref{sec:model_detail}. We demonstrate the out-of-domain generalisability of UniVTG-NA in Sec.~\ref{sec:generalisability}. We further motivate the need for negative-aware methods in video moment retrieval by displaying results with negative queries from UniMD~\cite{zeng2024unimd} in Sec.~\ref{sec:unimd}. Finally, we show more qualitative results from UniVTG-NA on QVHighights and Charades-STA in Sec.~\ref{sec:qual}.

\section{Dataset Information}
\label{sec:dataset}

\subsection{Out-of-Domain Negative Query Generation}
\label{subsec:neg_query_generation}
As mentioned in the main paper, the out-of-domain negative query sentences were generated using a large language model (LLM). Four broad scenarios were used as query topics, these were ``competitive sport'',  ``animal behaviour'', ``physics laboratory'', and ``mathematics class''.
In Table \ref{tab:subcategories}, the subtopics for each of these topics are listed. The prompt and specific LLM used for each topic is also displayed.
Prompts were empirically chosen to ensure the quality and diversity of the generated sentences. For example, for the ``animal behaviour'' topic, it was found that using scientific names improved upon these aspects, hence most of the subtopics are scientific names.
The four scenarios were chosen as they represent scenarios which are unlikely to be present within the QVHighlights and Charades-STA datasets, which cover news, vlogs and household actions. The choice of 4 broad scenarios helps to ensure that the OOD negatives remain OOD and do not accidentally produce false negatives. By using prompts specifically describing the actions as short, unique and varied, we are able to get a wider range of sentences without the language becoming too decorative. Sample sentences from each scenario are displayed in Table~\ref{tab:example_sen}.
We use the same set of OOD Negatives for both QVHighlights~\cite{lei2021qvhl} and Charades-STA~\cite{gao2017tall}.

\begin{table*}[t]
\begin{minipage}{\textwidth}
\centering
\resizebox{\columnwidth}{!}{
\begin{tabular}{|l|ll|lll|l|l|}
\hline
Topic         & \multicolumn{2}{c|}{Competitive Sport}                                                                                      & \multicolumn{3}{c|}{Animal Behaviour}                                                                                                                                & \multicolumn{1}{c|}{Physics Laboratory}                                                                                                                                   & \multicolumn{1}{c|}{Mathematics Class}                                                                                                                                      \\ \hline
Model         & \multicolumn{2}{l|}{Chat-GPT (GPT-4o)}                                                                                               & \multicolumn{3}{l|}{Claude 3 Opus}                                                                                                                                   & Claude 3 Opus                                                                                                                                                             & Claude 3 Opus                                                                                                                                                               \\ \hline
Prompt        & \multicolumn{2}{l|}{\begin{tabular}[c]{@{}l@{}}Generate X sentences describing\\ actions in $<$subtopic$>$\end{tabular}} & \multicolumn{3}{l|}{\begin{tabular}[c]{@{}l@{}}Generate $X$ unique and varied short sentences \\ of visual actions carried out by $<$subtopic$>$\end{tabular}} & \begin{tabular}[c]{@{}l@{}}Generate $X$ unique and\\ varied sentences of visual \\ actions carried out by a \\ person working in a \\ $<$subtopic$>$ lab.\end{tabular} & \begin{tabular}[c]{@{}l@{}}Generate $X$ unique and\\ varied sentences of visual \\ actions carried out by a \\ person working in a \\ $<$subtopic$>$ class.\end{tabular} \\ \hline
Subtopics & american football                                               & archery                                                   & accipitriformes                                        & agnatha                                                & alcidae                                            & acoustics                                                                                                                                                                 & algebra                                                                                                                                                                     \\
              & athletics field events                                          & badminton                                                 & anatidae                                               & anguilliformes                                         & annelids                                           & atmospheric physics                                                                                                                                                       & applied mathematics                                                                                                                                                         \\
              & baseball                                                        & boxing                                                    & anura                                                  & big cats                                               & bivalves                                           & biophysics                                                                                                                                                                & calculus                                                                                                                                                                    \\
              & cricket                                                         & cycling                                                   & bovidae                                                & camelid                                                & canidae                                            & chemical physics                                                                                                                                                          & combinatorics                                                                                                                                                               \\
              & darts                                                           & fencing                                                   & cephalopods                                            & cervidae                                               & chelicerata                                        & classical mechanics                                                                                                                                                       & computational maths                                                                                                                                                         \\
              & field hockey                                                    & golf                                                      & chiroptera                                             & chondrichthyes                                         & cnidaria                                           & condensed matter physics                                                                                                                                                  & geometry                                                                                                                                                                    \\
              & gymnastics                                                      & ice hockey                                                & crocodilia                                             & decapods                                               & echinoderms                                        & cosmology                                                                                                                                                                 & graph theory                                                                                                                                                                \\
              & ice skating                                                     & kickboxing                                                & elasmobranchs                                          & gastropods                                             & giraffidae                                         & electromagnetism                                                                                                                                                          & number theory                                                                                                                                                               \\
              & lacrosse                                                        & rowing                                                    & hymenoptera                                            & insects                                                & lagomorphs                                         & electronics                                                                                                                                                               & probability                                                                                                                                                                 \\
              & rugby                                                           & running                                                   & lepidoptera                                            & lizards                                                & marsupials                                         & fluid dynamics                                                                                                                                                            & statistics                                                                                                                                                                  \\
              & skateboarding                                                   & skiing                                                    & monotremes                                             & mustelids                                              & osteichthyes                                       & geophysics                                                                                                                                                                &                                                                                                                                                                             \\
              & snooker                                                         & snowboarding                                              & pinnipeds                                              & platyhelminthes                                        & porifera                                           & medical physics                                                                                                                                                           &                                                                                                                                                                             \\
              & soccer                                                          & squash                                                    & primates                                               & proboscidea                                            & ratites                                            & optical physics                                                                                                                                                           &                                                                                                                                                                             \\
              & swimming                                                        & table tennis                                              & rodents                                                & serpentes                                              & spheniscidae                                       & particle physics                                                                                                                                                          &                                                                                                                                                                             \\
              & tennis                                                          & ultimate frisbee                                          & stomatopods                                            & strigiformes                                           & suina                                              & quantum mechanics                                                                                                                                                         &                                                                                                                                                                             \\
              & water polo                                                      &                                                           & talpidae                                               & testudines                                             & urodela                                            & thermodynamics                                                                                                                                                            &                                                                                                                                                                             \\
              &                                                                 &                                                           & ursidae                                                & wading birds                                           &                                                    &                                                                                                                                                                           &                                                                                                                                                                             \\ \hline
\end{tabular}}
\caption{List of topics and subtopics used for out-of-domain negative generation, along with the prompts and LLMs used. $X$ represents the number of sentences requested which varied from 50 to 100.}
\label{tab:subcategories}
\end{minipage}\hfill
\end{table*}

\begin{table*}[t]
\begin{minipage}{\textwidth}
\centering
\resizebox{\columnwidth}{!}{
\begin{tabular}{|ll|}
\hline
\multicolumn{1}{|c|}{Competitive Sport}                                                                                                                       & \multicolumn{1}{c|}{Animal Behaviour}                                                                                        \\ \hline
\multicolumn{1}{|l|}{The outfielder throws to home plate}                                                                                                     & An osprey dives into the water, snatching a fish with its talons                                                             \\
\multicolumn{1}{|l|}{The opponent hits a drop shot followed by a lob}                                                                                         & A northern harrier glides low over a meadow, searching for small mammals.                                                    \\
\multicolumn{1}{|l|}{The striker heads the ball into the net}                                                                                                 & An ovambo sparrowhawk sits near its nest, guarding its eggs.                                                                 \\
\multicolumn{1}{|l|}{The punter pins the opposing team deep in their own territory with a well-placed kick}                                                   & A slender-billed kite hunts for insects over an African grassland                                                            \\
\multicolumn{1}{|l|}{The player executes a deceptive backhand drop shot}                                                                                      & A white-backed vulture strips meat from a carcass with its strong beak                                                       \\
\multicolumn{1}{|l|}{The opponent flicks a shuttlecock deep into the backcourt}                                                                               & A lamprey swam in a figure-eight pattern, leaving pheromone trails for potential mates                                       \\
\multicolumn{1}{|l|}{The goalie makes a sprawling save}                                                                                                       & A group of lamprey larvae anchored themselves to rocks, facing into the current                                              \\
\multicolumn{1}{|l|}{The player switches to a colored ball after potting all reds}                                                                            & A bronze eel lay coiled on the seafloor, its coppery scales gleaming                                                         \\
\multicolumn{1}{|l|}{The opponent covers up, absorbing the blows}                                                                                             & A topaz eel darted through a school of yellow tang, its golden body mirroring their color                                    \\
\multicolumn{1}{|l|}{The center offloads the ball to a teammate before being tackled}                                                                         & A bootlace worm tangles itself around a piece of driftwood                                                                   \\
\multicolumn{1}{|l|}{The flanker disrupts the opposing team's maul, forcing a turnover}                                                                       & A red-eyed tree frog clings to a leaf with its sticky toe pads                                                               \\
\multicolumn{1}{|l|}{The rowers maintain their balance as the boat rocks gently on the water}                                                                 & The Amazon milk frog inflated its body, trying to appear larger                                                              \\
\multicolumn{1}{|l|}{The rowers return to the dock and disembark from the boat}                                                                               & The jaguar's powerful jaws crushed the turtle's shell                                                                        \\
\multicolumn{1}{|l|}{The archer aims downrange, focusing on the target}                                                                                       & A long clam extends its siphons, drawing in water to filter out food particles                                               \\
\multicolumn{1}{|l|}{The skater lands a double axel with precision}                                                                                           & A kudu reached up to browse on acacia tree leaves                                                                            \\
\multicolumn{1}{|l|}{The swimmer's streamline position reduces resistance through the water}                                                                  & A Pale fox kit playfully wrestles with its sibling outside their den                                                         \\
\multicolumn{1}{|l|}{Skiers maintain a tight tuck to minimize drag}                                                                                           & A Cape fox, known for its nocturnal habits, emerges from its den at dusk to begin hunting                                    \\
\multicolumn{1}{|l|}{The athlete lands on the mat on the other side of the bar}                                                                               & The moose browsed on the tender bark of a young tree, stripping it with its teeth                                            \\
\multicolumn{1}{|l|}{The skater executes a jump combination, linking jumps of different rotations}                                                            & A crab spider ambushed a bee from its hiding spot in a flower                                                                \\
\multicolumn{1}{|l|}{The gymnast tumbles with precision on the floor exercise mat}                                                                            & A moon coral's large, rounded polyps resemble a cluster of full moons                                                        \\ \hline
\multicolumn{1}{|c|}{Physics Laboratory}                                                                                                                      & \multicolumn{1}{c|}{Mathematics Class}                                                                                       \\ \hline
\multicolumn{1}{|l|}{The researcher measured the sound absorption coefficient of the new acoustic material}                                                   & They create a flow diagram to show the steps in the algorithm                                                                \\
\multicolumn{1}{|l|}{The acoustician measured the sound reduction index of the window using a pink noise generator}                                           & He draws a box-and-whisker plot to compare the distributions of different data sets                                          \\
\multicolumn{1}{|l|}{He used a sound intensity probe to measure the sound power of the jet engine}                                                            & They create a Venn diagram to find the probability of the union of two events                                                \\
\multicolumn{1}{|l|}{The researcher uses a ceilometer to determine the height of the cloud base}                                                              & She uses the separation of variables technique to solve the partial differential equation                                    \\
\multicolumn{1}{|l|}{With a steady hand, the chemist uses a capillary tube to load the viscous ionic liquid into the rheometer for flow behavior studies}     & He arranges a set of numbered tiles to illustrate the concept of permutations with repetition                                \\
\multicolumn{1}{|l|}{The graduate student intently studies the XPS spectrum, identifying the chemical states of the elements present on the catalyst surface} & With a critical eye, she examined the partial dependence plots, assessing the impact of individual features on the model     \\
\multicolumn{1}{|l|}{She recorded the data from the oscilloscope in her lab notebook}                                                                         & He labels each vertex with a unique letter, making it easier to refer to specific nodes                                      \\
\multicolumn{1}{|l|}{The scientist replaced the filament in the electron gun}                                                                                 & She shades a vertex to indicate it has been visited during a graph traversal                                                 \\
\multicolumn{1}{|l|}{He carefully positioned the sample in the center of the split-coil magnet}                                                               & He draws a graph with a minimum spanning tree, a subgraph connecting all vertices with the minimum total edge weight         \\
\multicolumn{1}{|l|}{The scientist adjusted the settings on the surface plasmon resonance (SPR) instrument}                                                   & He shades the area representing the union of two probability events                                                          \\
\multicolumn{1}{|l|}{The cosmologist carefully positioned the spectrograph, ready to analyze the light from a distant supernova}                              & The statistician carefully folded the large printed graph, ensuring the creases were sharp and the edges aligned             \\
\multicolumn{1}{|l|}{He studies the flow patterns in a porous medium using magnetic resonance imaging}                                                        & The analyst used a highlighter to trace the trend line on the time series plot                                               \\
\multicolumn{1}{|l|}{The researcher measures the thermal conductivity of a rock sample using a divided bar apparatus}                                         & The statistician used a chalk line to draw a perfectly straight line on the chalkboard, representing the regression equation \\
\multicolumn{1}{|l|}{The geophysicist uses a Schmidt hammer to test the strength of a rock outcrop}                                                           & He leans forward, listening intently to his colleague's explanation of a new mathematical technique                          \\
\multicolumn{1}{|l|}{The physicist calibrated the radionuclide calibrator for accurate activity measurements of radiopharmaceuticals}                         & The mathematician creates a Pascal's triangle, highlighting the connection between combinatorics and binomial coefficients   \\
\multicolumn{1}{|l|}{The scientist adjusted the position of the camera to capture the desired image}                                                          & She writes out the formula for calculating the number of combinations of n objects taken r at a time                         \\
\multicolumn{1}{|l|}{He measured the wavelength of the light using a spectrometer}                                                                            & He creates a matrix to represent the adjacency relationships in a combinatorial graph                                        \\
\multicolumn{1}{|l|}{She calculates the probability of a defective product using quality control data}                                                        & He arranges a set of dominoes in different configurations, exploring the number of possible tilings                          \\
\multicolumn{1}{|l|}{She adjusts the phase shifter to control the interference between the microwave signals}                                                 & The mathematician draws a tree diagram to illustrate the Collatz conjecture                                                  \\
\multicolumn{1}{|l|}{The scientist uses a laser thermometer to measure the surface temperature of the material}                                               & She writes out a proof using mathematical induction, establishing a pattern                                                  \\ \hline
\multicolumn{2}{|c|}{Musician Performance}                                                                                                                                                                                                                                                   \\ \hline
\multicolumn{2}{|l|}{The accordionist's fingers danced across the keys, effortlessly transitioning between notes}                                                                                                                                                                            \\
\multicolumn{2}{|l|}{The man blows into the blowpipe to fill the bag with air}                                                                                                                                                                                                               \\
\multicolumn{2}{|l|}{The banjo player's hands moved in a blur, creating an intricate fingerpicking pattern}                                                                                                                                                                                  \\
\multicolumn{2}{|l|}{He muted the strings with his palm, creating a staccato effect}                                                                                                                                                                                                         \\
\multicolumn{2}{|l|}{The bongo player's hands alternate between drums}                                                                                                                                                                                                                       \\
\multicolumn{2}{|l|}{The cellist leans into the instrument, conveying the emotion of the piece through their posture}                                                                                                                                                                        \\
\multicolumn{2}{|l|}{She brushes the snare drum lightly, creating a soft, sizzling sound}                                                                                                                                                                                                    \\
\multicolumn{2}{|l|}{Their cheeks puff out as they blow into the mouthpiece}                                                                                                                                                                                                                 \\
\multicolumn{2}{|l|}{She plays the guitar while sitting on a stool}                                                                                                                                                                                                                          \\
\multicolumn{2}{|l|}{He tapped his fingers on the fretboard, creating a percussive rhythm}                                                                                                                                                                                                   \\
\multicolumn{2}{|l|}{He alternates between blowing and drawing on the harmonica, creating a dynamic sound}                                                                                                                                                                                   \\
\multicolumn{2}{|l|}{With closed eyes, the musician swayed gently as they strummed the harp's delicate strings}                                                                                                                                                                              \\
\multicolumn{2}{|l|}{She gently presses the white keys with her fingertips}                                                                                                                                                                                                                  \\
\multicolumn{2}{|l|}{She places her feet on the pedals and her hands on the keys}                                                                                                                                                                                                            \\
\multicolumn{2}{|l|}{She smiled at the audience, her saxophone gleaming under the stage lights as she played a upbeat tune}                                                                                                                                                                  \\
\multicolumn{2}{|l|}{He slides his left hand along the strings to change the pitch}                                                                                                                                                                                                          \\
\multicolumn{2}{|l|}{They play a glissando by sliding their finger across the keys}                                                                                                                                                                                                          \\
\multicolumn{2}{|l|}{They keep their hands steady for a long, sustained note}                                                                                                                                                                                                                \\
\multicolumn{2}{|l|}{He tilts the trombone up for a high note}                                                                                                                                                                                                                               \\
\multicolumn{2}{|l|}{The musician's eyes darted between the sheet music and his fingers, ensuring he played each note correctly}                                                                                                                                                             \\ \hline
\end{tabular}}
\caption{Example sentences from each OOD topic.}
\label{tab:example_sen}
\end{minipage}\hfill
\end{table*}

\subsection{Negative Aware Dataset Details}

Table \ref{tab:data_numbers} displays the number of positive and negative queries used during training/evaluation of the models. For Charades-STA, where there are fewer negative queries than positive for out-of-domain, the negative queries are assigned to multiple videos. This still produces a distinct signal as each video-sentence pair offers a different semantic relationship.

\begin{table*}[t]
\centering
\begin{tabular}{llllllll}
\hline
             & \multicolumn{3}{c}{Train}                                                                                                                    &  & \multicolumn{3}{c}{Test}                                                                                                                 \\ \cline{2-4} \cline{6-8} 
             & Positive & \begin{tabular}[c]{@{}l@{}}In-Domain \\ Negative\end{tabular} & \begin{tabular}[c]{@{}l@{}}Out-of-Domain \\ Negative\end{tabular} &  & Positive & \begin{tabular}[c]{@{}l@{}}In-Domain \\ Negative\end{tabular} & \begin{tabular}[c]{@{}l@{}}Out-of-Domain \\ Negative\end{tabular} \\ \hline
QVHighlights~\cite{lei2021qvhl} & 7218     & 7218                                                          & 7230                                                              &  & 1550     & 1550                                                          & 1550                                                              \\
Charades-STA~\cite{gao2017tall} & 12404    & 12404                                                         & 7230                                                              &  & 3720     & 3720                                                          & 1550                                                              \\ \hline
\end{tabular}
\caption{Numbers of positive and negative queries used for QVHighlights and Charades-STA.}
\label{tab:data_numbers}
\end{table*}

\section{SVM Trained on Saliency Scores}
\label{sec:svms}
We train an SVM on the outputted positive and negative query saliency scores from UniVTG~\cite{lin2023univtg}, QD-DETR~\cite{moon2023query} and CG-DETR~\cite{moon2024cgdetr} for the QVHighlights and Charades-STA datasets. This is to quantify how separable positive and negative queries are when the relationship between them is modelled using saliency outputs from the base models, without any explicit training for negative rejection.
Results are displayed in Table~\ref{tab:svm}.

\begin{table*}[ht]
\begin{minipage}{\textwidth}
    \centering
    \resizebox{\columnwidth}{!}{
    \begin{tabular}{lccrrlccrr}
    \toprule
                &\multicolumn{4}{c}{QVHighlights}&&\multicolumn{4}{c}{Charades-STA}\\ \cmidrule{2-5} \cmidrule{7-10}
                &     &     &\multicolumn{2}{c}{Rejection Acc. (\%)}&&     &     &\multicolumn{2}{c}{Rejection Acc. (\%)}\\ \cmidrule{4-5} \cmidrule{9-10}
    Method      &R1@0.5&R1@0.7&\multicolumn{1}{c}{ID}&\multicolumn{1}{c}{OOD}&&R1@0.5&R1@0.7&\multicolumn{1}{c}{ID}&\multicolumn{1}{c}{OOD}\\ \midrule
    UniVTG~\cite{lin2023univtg} SVM      &63.48 (\textcolor{red}{-3.87})&49.87 (\textcolor{red}{-2.78})&  94.77            &  97.74         &&53.47 (\textcolor{red}{-6.75})&33.49 (\textcolor{red}{-5.06})& 35.89             & 50.40          \\
    CG-DETR~\cite{moon2024cgdetr} SVM  &62.84 (\textcolor{red}{-4.26}) & 50.26 (\textcolor{red}{-3.29})& 95.55 & 95.41         &&43.79 (\textcolor{red}{-13.74})&28.44 (\textcolor{red}{-7.23})& 82.90             & 74.52          \\
    QD-DETR~\cite{moon2023query} SVM &48.26 (\textcolor{red}{-13.74}) & 37.42 (\textcolor{red}{-8.84}) & 96.32 & 95.09         &&46.67 (\textcolor{red}{-12.44}) & 30.05 (\textcolor{red}{-6.70}) & 76.91 & 81.59          \\
    \bottomrule
    \end{tabular}}
    \caption{Results of training an SVM on top of the saliency score outputs of UniVTG, CG-DETR and QD-DETR.}
    \label{tab:svm}
    \end{minipage}\hfill
\end{table*}

The SVM results on QVHighlights show high rejection accuracy at the cost of decreased R1@$\theta$ scores. In the case of QD-DETR, these are significantly decreased. The Charades-STA results show reasonable rejection accuracy at significant cost to the R1@$\theta$ scores for CG-DETR and QD-DETR, while UniVTG fails to achieve high rejection accuracy but has better R1@$\theta$ scores. Overall these results display the limitations of using the saliency outputs from the base models alone for combined moment retrieval and negative rejection, particularly on datasets without ground-truth saliency scores such as Charades-STA. It further motivates the need to train explicitly for negative rejection.

\section{Model Details}
\label{sec:model_detail}
\subsection{UniVTG-NA}
\label{sec:univtg}
For UniVTG-NA, the input to the classification head is a direct sum of the indicator scores and saliency scores. \ie $g_i = f_i + s_i$ where $g_i$ is the classification head input at index $i$. 

\paragraph{Loss Adaptations.}
We specify the adjustments made to the losses for the UniVTG-NA model from UniVTG~\cite{lin2023univtg}. Aside from the boundary prediction losses being set to 0 for the negative queries, the saliency losses are also adjusted. UniVTG uses a saliency loss $\mathcal{L}_s$ which is a weighted summation of inter-video and intra-video contrastive losses. It is not possible to use the contrastive saliency loss with negative queries. Therefore, for negative queries the saliency loss is defined as a loss applied directly on the cosine similarity between the video clip $\textbf{v}_i$, and sentence features $\textbf{S}$, with $\lambda_s^-$ as the loss weighting.
\begin{equation}
   \mathcal{L}_s^- = \lambda_s^-\cos(\textbf{v}_i, \textbf{S}) := \lambda_s^-\frac{\textbf{v}_i^T\textbf{S}}{\lVert\textbf{v}_i\rVert_2\lVert\textbf{S}\rVert_2}
   \label{eq:cosine}
\end{equation}
This is done as the saliency scores are computed via cosine similarity between sentence and video clip features for UniVTG, so achieves our principle of designing the saliency loss for negatives such that it pushes the saliency scores lower.
Furthermore, for UniVTG-NA's foreground matching loss with negative queries, $\mathcal{L}_f^- = \mathcal{L}_f$ as no adjustments are made to the matching loss, which is a BCE loss on the individual indicator scores. This already achieves the aim of pushing the indicator scores lower.

\subsection{QD-DETR-NA \& CG-DETR}
\label{sec:qddetr}
\begin{figure}[t]
    \centering
    \includegraphics[width = \linewidth]{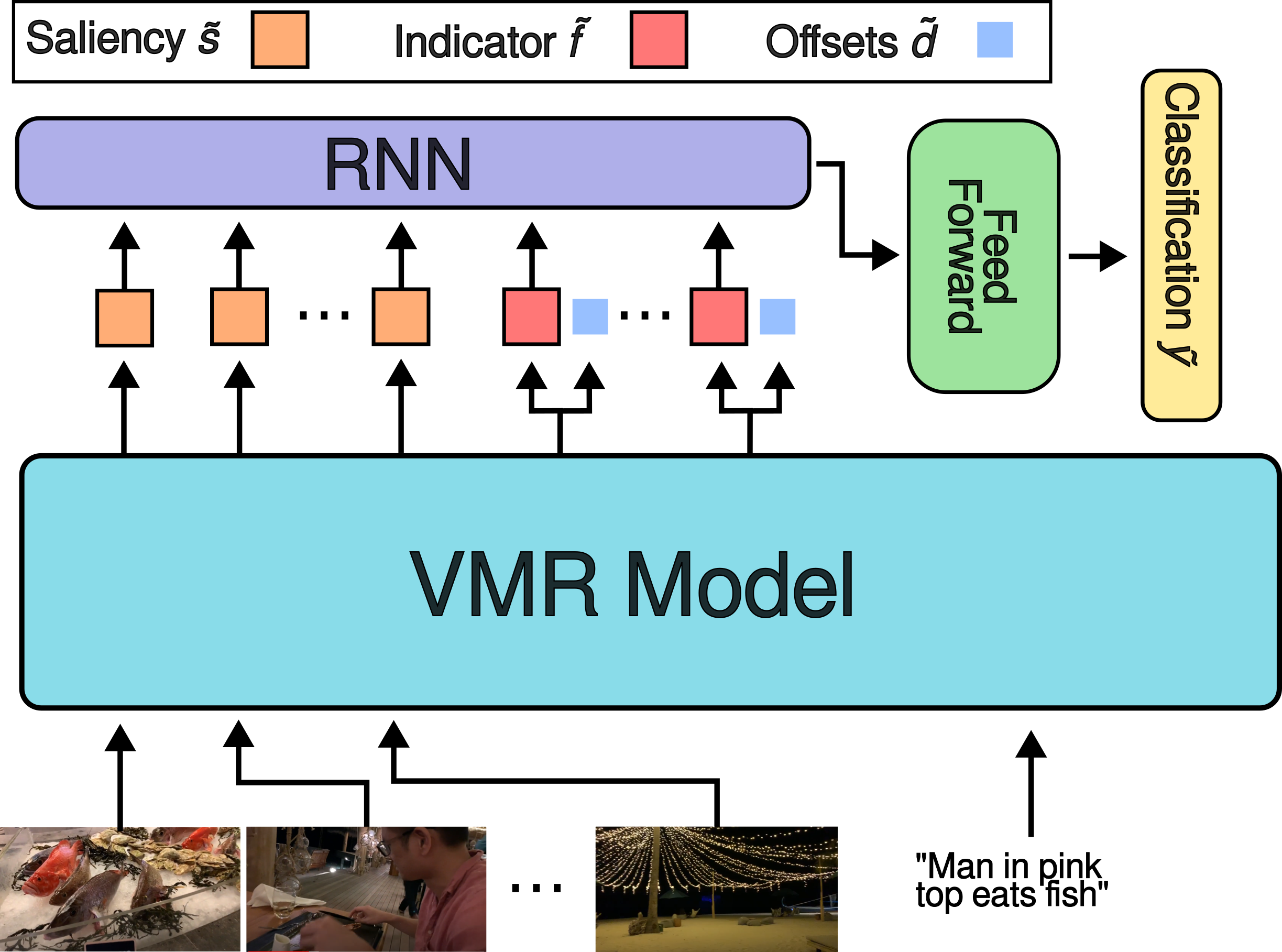}
    \caption{The classification head for QD-DETR-NA and CG-DETR-NA takes as input a concatenation of saliency and indicator scores, which are then passed through a recurrent layer and a feed forward layer before producing a single value output for classification.}
    \label{fig:method_qd_detr}
        \vspace*{-12pt}
\end{figure}

Negative-aware versions of QD-DETR and CG-DETR were also trained to evaluate the proposed method on other models. The details of the QD-DETR and CG-DETR implementation are as follows:
Given the indicator score outputs $\{\tilde{f}_1,...,\tilde{f}_M\}$ and saliency score outputs $\{\tilde{s}_1,...,\tilde{s}_{L_v}\}$, the input to the classification head is a concatenation $g = \{\tilde{s}_1,...,\tilde{s}_{L_v}, \tilde{f}_1,...,\tilde{f}_M\} \in \mathbb{R}^{(L_v+M)}$, where $L_v$ is the number of video clip features and $M$ is the number of moment queries. This implementation is represented in Figure~\ref{fig:method_qd_detr}. This is chosen as opposed to a summation because QD-DETR/CG-DETR use moment queries to generate the moment candidates rather than just the text-attended video clip representations from the encoder. In this case, there is not a one-to-one correspondence with the saliency scores, \ie $L_v \neq M$. 

As with UniVTG, the boundary losses were set to 0 and the foreground matching loss was retained for the negative queries. For both methods, the saliency loss has three components, two of which are contrastive and are therefore not feasible for negative queries. The remaining loss works to reduce the negative query saliency scores, thus achieving the principle aim of the negative query saliency loss. Therefore it is retained as the sole loss for negative queries. It is shown for a saliency score output $s_i$ with loss weighting $\lambda^-_s$ below.
\begin{equation}
    \mathcal{L}_s^- = \lambda^-_s(-\log(1-s_i))
    \label{eq:sal_loss}
\end{equation}

\subsection{Implementation Details}
\label{subsec:implementation_details}
\paragraph{UniVTG}
For QVHighlights, we use loss weightings of $\lambda^+ = 1$, $\lambda^-_{ID} = 0.1$, $\lambda^-_{OOD} = 0.1$, and $\lambda_p = 1$, while for Charades-STA, we adjust $\lambda^-_{ID} = 0.5$, $\lambda^-_{OOD} = 0.5$. The remaining loss weightings are retained from QVHighlights and Charades-STA training defaults in UniVTG. For negative queries the cosine similarity loss weighting $\lambda_s^-$ is set equal to the intra video saliency loss weighting.
\paragraph{QD-DETR}
Loss weightings of $\lambda^+ = 1$, $\lambda^-_{ID} = 0.05$, $\lambda^-_{OOD} = 0.05$, and $\lambda_p = 1$ are used for both QVHighlights and Charades-STA. For QVHighlights, $\lambda^-_s = 1$ while for Charades-STA, $\lambda^-_s = 4$.
\paragraph{CG-DETR}
The same weightings are used as in QD-DETR except $\lambda^-_{ID} = 0.1$, $\lambda^-_{OOD} = 0.1$ for both datasets. All other loss weightings retain their default values.

\section{OOD Generalisability}
\label{sec:generalisability}
To test the generalisability of the negative-aware approach for OOD query sentences, we test the UniVTG-NA model on OOD sentences from another scenario on which the model has not been trained. This scenario is `musician performances' (see sample sentences in Table \ref{tab:example_sen}). The rejection accuracy results are shown in Table \ref{tab:music}. The rejection accuracy remains high for both datasets, demonstrating that the model is capable of generalising to other OOD scenarios.

\begin{table}[ht!]
\begin{tabular}{lcc}
\toprule
          & \multicolumn{2}{c}{Rejection Acc. (\%)}                     \\ \cline{2-3} 
Method    & QVHighlights             & \multicolumn{1}{c}{Charades-STA} \\ \hline
UniVTG-NA & \multicolumn{1}{c}{99.8} & 93.8                             \\ \bottomrule
\end{tabular}
\caption{Rejection accuracy results for UniVTG-NA on the unseen OOD category of `musician performance'.}
\label{tab:music}
\end{table}

\section{UniMD}
\label{sec:unimd}
\begin{figure*}[t]
    \centering
    \includegraphics[width = \linewidth]{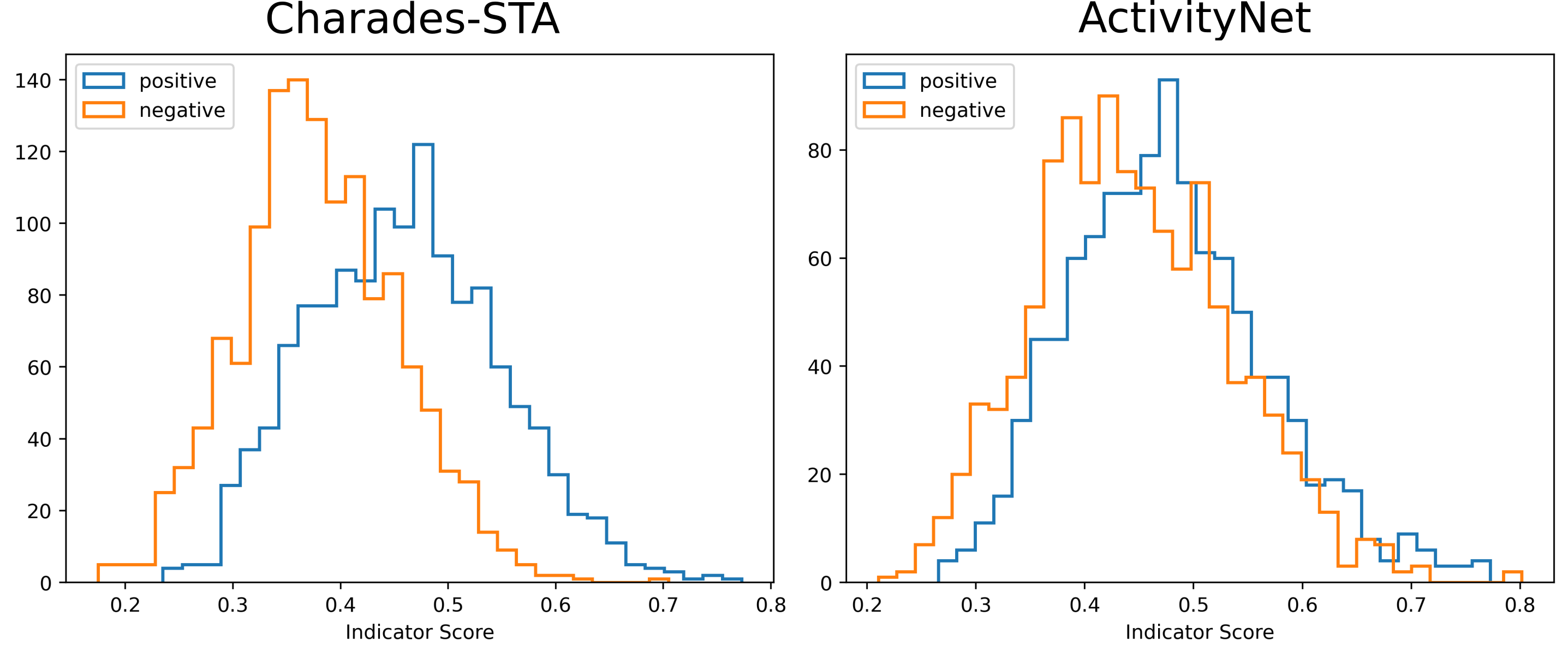}
    \caption{Histograms of prediction (indicator) scores for positive and in-domain negative queries produced by the UniMD model.}
    \label{fig:histogram}
\end{figure*}

To further motivate the need for negative-aware training for the task of Negative-Aware Video Moment Retrieval, we investigate the output produced by UniMD~\cite{zeng2024unimd}, a recent SOTA method which only produces indicator scores with no saliency scores. We plot histograms of the output scores for positive and in-domain negative sentences for Charades-STA and ActivityNet-Captions, as in Figure \ref{fig:histogram}. There is significant overlap between the positive and negative distributions which shows that the model is not designed to handle negative rejection. This further motivates the need for models which are specifically trained to carry out negative rejection alongside moment retrieval.

\section{Qualitative Results}
\label{sec:qual}
\begin{figure*}[t]
    \centering
    \includegraphics[width=0.99\linewidth]{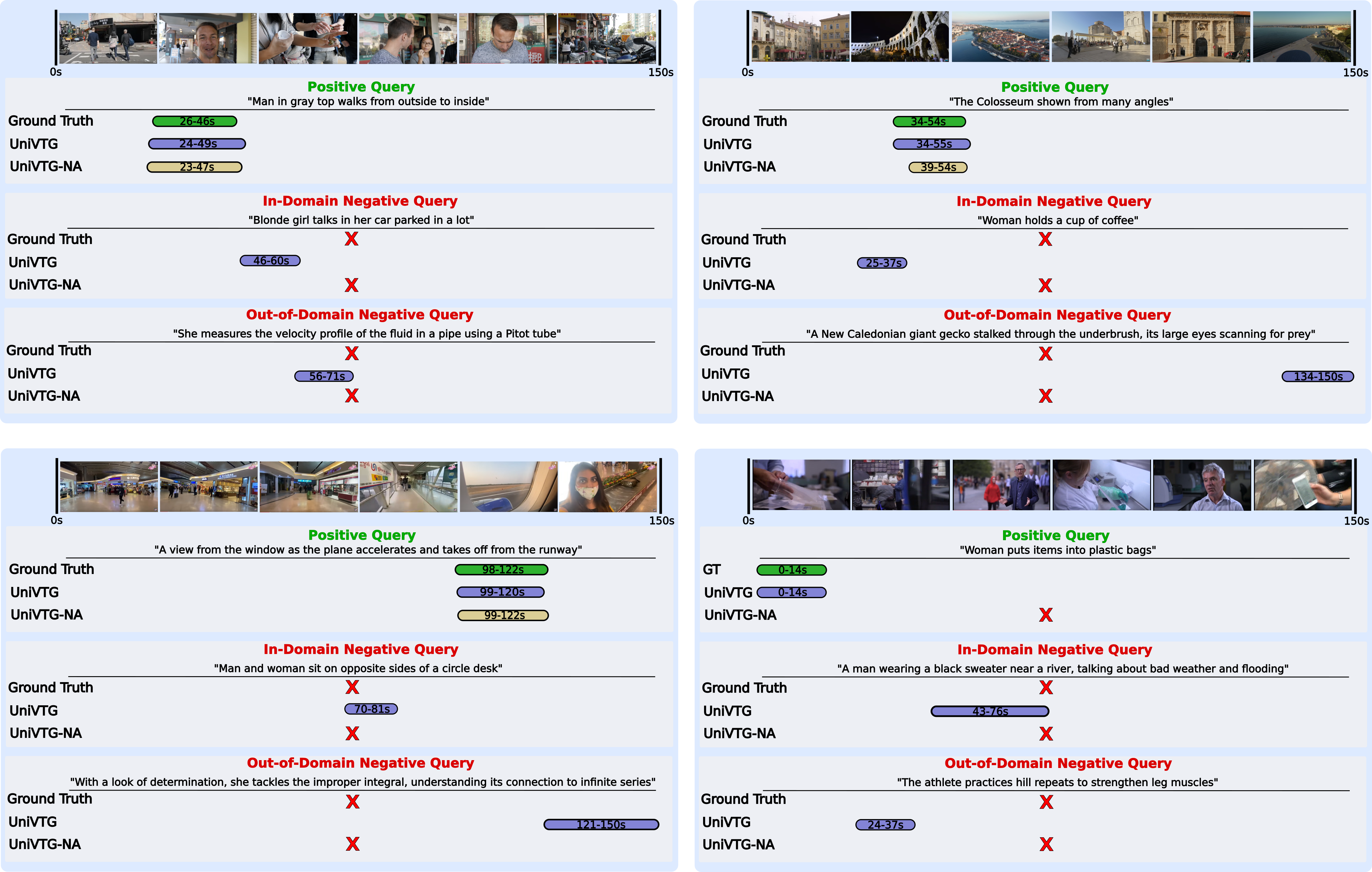}
    \caption{Qualitative results from UniVTG-NA on QVHighlights.}
    \label{fig:qual_qvhl}
\end{figure*}

\begin{figure*}[t]
    \centering
    \includegraphics[width=0.99\linewidth]{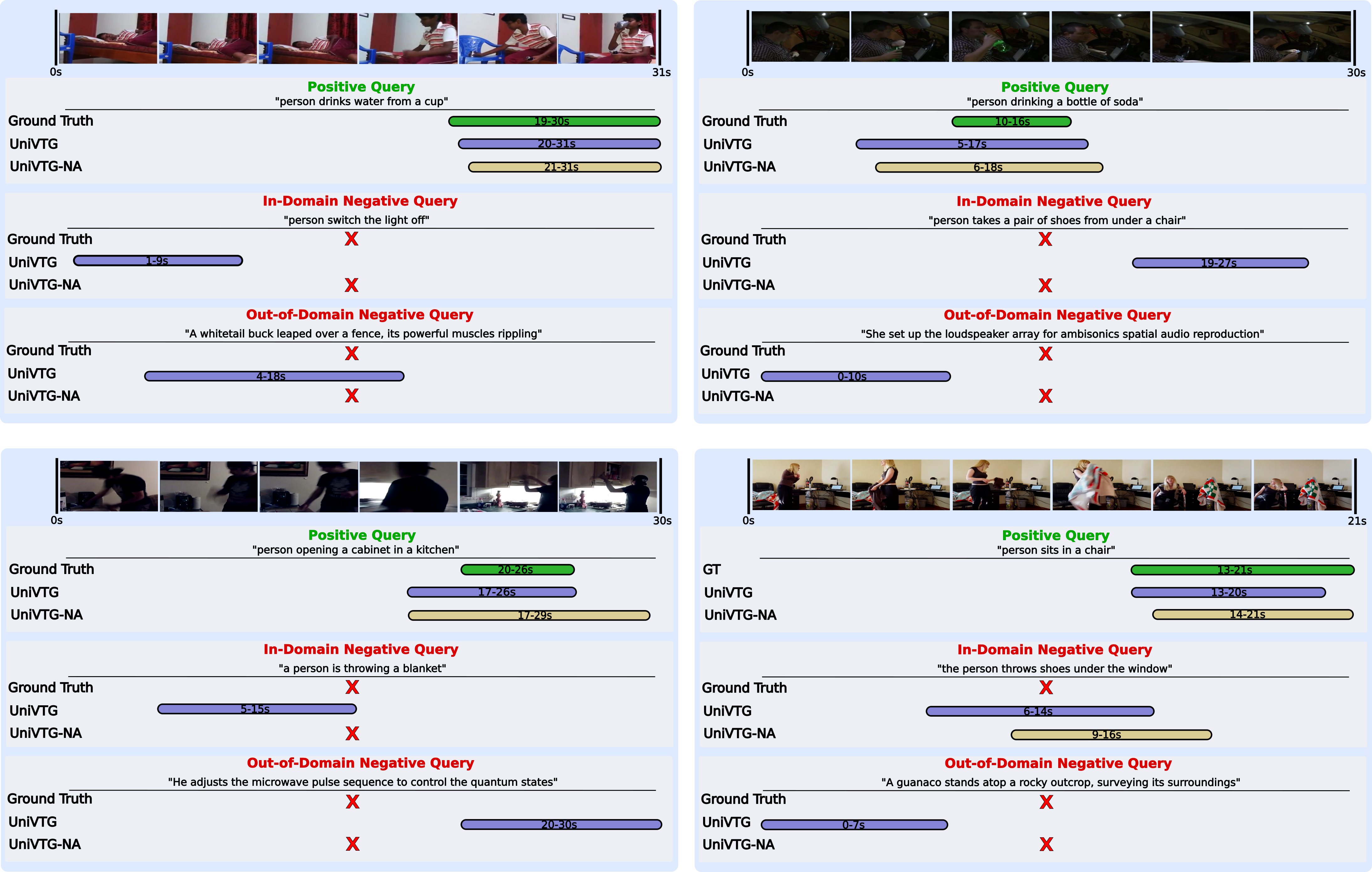}
    \caption{Qualitative results from UniVTG-NA on Charades-STA.}
    \label{fig:qual_charades}
        \vspace*{-12pt}
\end{figure*}

We provide further qualitative results from UniVTG-NA on the QVHighlights and Charades-STA datasets in Figure \ref{fig:qual_qvhl} \& \ref{fig:qual_charades}. The model frequently successfully localises the positive sentences and rejects the negative sentences. Failure cases are included in the bottom right of each set of examples. The failure case in Figure \ref{fig:qual_qvhl} is a case of UniVTG-NA rejecting a positive sentence, while in Figure \ref{fig:qual_charades} the model fails to reject an ID negative sentence.

\end{document}